\documentclass[10pt,twocolumn,letterpaper]{article}

\usepackage{cvpr}
\usepackage{times}
\usepackage{epsfig}
\usepackage{graphicx}
\usepackage{amsmath}
\usepackage{amssymb}
\usepackage[noend]{algpseudocode}
\usepackage{algorithm}

\usepackage{multirow}
\usepackage{mathtools}
\usepackage{siunitx}
\usepackage{afterpage}
\DeclarePairedDelimiter{\ceil}{\lceil}{\rceil}

\usepackage[pagebackref=true,breaklinks=true,letterpaper=true,colorlinks,bookmarks=false]{hyperref}

\cvprfinalcopy 

\def\cvprPaperID{2464} 

\ifcvprfinal\fi
\begin{document}

\title{Fast Algorithms for Convolutional Neural Networks}

\author{Andrew Lavin\\
{\tt\small alavin@acm.org}
\and
Scott Gray\\
Nervana Systems\\
{\tt\small sgray@nervanasys.com}
}

\maketitle

\begin{abstract}
Deep convolutional neural networks take GPU days of compute time to train on large data sets. Pedestrian detection  for self driving cars requires very low latency. Image recognition for mobile phones is constrained by limited processing resources. The success of convolutional neural networks in these situations is limited by how fast we can compute them. Conventional FFT based convolution is fast for large filters, but state of the art convolutional neural networks use small, $3 \times 3$ filters. We introduce a new class of fast algorithms for convolutional neural networks using Winograd's minimal filtering algorithms. The algorithms compute minimal complexity convolution over  small tiles, which makes them fast with small filters and small batch sizes. We benchmark a GPU implementation of our algorithm with the VGG network and show state of the art throughput at batch sizes from 1 to 64. 
\end{abstract}

\section{Introduction}

Deep convolutional neural networks (convnets) achieve state of the art results on image recognition problems~\cite{simonyan2014very}\cite{ioffe2015batch}. The networks take several days of GPU time to train and require significant compute resources during classification as well. Larger data sets and models lead to better accuracy but also increase computation time. Therefore progress in deep neural networks is limited by how fast the networks can be computed.

Likewise the application of convnets to low latency inference problems, such as pedestrian detection in self driving car video imagery, is limited by how fast a small set of images, possibly a single image, can be classified.

Distributed training of convnets can be achieved by partitioning each batch of examples across the nodes of a cluster and accumulating weight updates across the nodes. Large batch sizes adversely affect convergence of the network, so the minimum batch size that can be computed efficiently places an upper limit on cluster size~\cite{krizhevsky2014one,gupta2015model}.

State of the art convnet architectures for image recognition use deep networks of $3\times 3$ convolutional layers, because they achieve better accuracy with fewer weights than shallow networks with larger filters~\cite{simonyan2014very,ioffe2015batch}.

Therefore there is a strong need for fast convnet algorithms for small batch sizes and small filters. However conventional convnet libraries require large batch sizes and large filters for fast operation.

This paper introduces a new class of fast algorithms for convolutional neural networks based on the minimal filtering algorithms pioneered by Winograd~\cite{winograd1980arithmetic}. The algorithms can reduce the arithmetic complexity of a convnet layer by up to a factor of 4 compared to direct convolution. Almost all of the arithmetic is performed by dense matrix multiplies of sufficient dimensions to be computed efficiently, even when the batch size is very small. The memory requirements are also light compared to the conventional FFT convolution algorithm. These factors make practical implementations possible. Our implementation for NVIDIA Maxwell GPUs achieves state of the art throughput for all batch sizes measured, from 1 to 64, while using at most 16MB of workspace memory.

\section{Related Work}

The FFT and convolution theorem have been used to reduce the arithmetic complexity of convnet layers, first by Mathieu \etal~\cite{DBLP:journals/corr/MathieuHL13}, then refined by Visalache \etal~\cite{DBLP:journals/corr/VasilacheJMCPL14}, and then implemented in the NVIDIA cuDNN library~\cite{cuDNN}.

The Strassen algorithm for fast matrix multiplication was used by Cong and Xiao~\cite{cong2014minimizing} to reduce the number of convolutions in a convnet layer, thereby reducing its total arithmetic complexity. The authors also suggested that more techniques from arithmetic complexity theory might be applicable to convnets.

Various approaches have been attempted to reduce the complexity of convnets by quantizing or otherwise approximating the convolutional layer. We consider these approaches as orthogonal and complementary to those that exploit algebraic structure, and therefore declare them outside the scope of this paper.

\section{Convolutional Neural Networks}

A convnet layer correlates a bank of $K$ filters with $C$ channels and size $R \times S$ against a minibatch of $N$ images with $C$ channels and size $H \times W$. We denote filter elements as $G_{k,c,u,v}$ and image elements as $D_{i,c,x,y}$.

The computation of a single covnnet layer output $Y_{i,k,x,y}$ is given by the formula:
\begin{equation}\label{eq:convnetdirect}
Y_{i,k,x,y} = \sum\limits_{c=1}^{C} \sum\limits_{v=1}^{R} \sum\limits_{u=1}^{S} D_{i,c,x+u,y+v} G_{k,c,u,v}
\end{equation}
and we can write the output of an entire image/filter pair as
\begin{equation}\label{eq:convnetcorrelation}
Y_{i,k} = \sum\limits_{c=1}^{C} D_{i,c} \ast G_{k,c}
\end{equation}
where $\ast$ denotes 2D correlation.

\section{Fast Algorithms}

It has been known since at least 1980 that the minimal filtering algorithm for computing $m$ outputs with an $r$-tap FIR filter, which we call $F(m,r)$, requires

\begin{equation}
\mu(F(m,r)) = m+r-1
\end{equation}
multiplications~\cite[p.~39]{winograd1980arithmetic}. Also, we can nest minimal 1D algorithms $F(m,r)$ and $F(n,s)$ to form minimal 2D algorithms for computing $m\times n$ outputs with an $r \times s$ filter, which we call $F(m \times n, r \times s)$. These require
\begin{equation}
\begin{split}
\mu (F(m \times n,r \times s)) &= \mu(F(m,r)) \mu(F(n,s)) \\
&= (m+r-1)(n+s-1)
\end{split}
\end{equation}
multiplications~\cite{winograd1980multiplication}. We can continue to nest 1D algorithms to form algorithms for multi-dimensional FIR filters.

It is interesting to note that in 1D, 2D, and multi-dimensions, the minimal algorithm requires a number of multiplications equal to the number of inputs. In other words, to compute $F(m,r)$ we must access an interval  of $m+r-1$ data values, and to compute $F(m \times n, r \times s)$ we must access a tile of $(m+r-1) \times (n+s-1)$ data values. Therefore the minimal filtering algorithm requires one multiplication per input.

\subsection{F(2x2,3x3)}

The standard algorithm for $F(2,3)$ uses $2 \times 3 = 6$ multiplications. Winograd~\cite[p.~43]{winograd1980arithmetic} documented the following minimal algorithm:
\begin{equation} \label{eq:F_2_3}
F(2,3) = \begin{bmatrix}
d_0 && d_1 && d_2\\
d_1 && d_2 && d_3\\
\end{bmatrix}
\begin{bmatrix}
g_0\\
g_1\\
g_2\\
\end{bmatrix}
=
\begin{bmatrix}
m_1+m_2+m_3\\
m_2-m_3-m_4\\
\end{bmatrix}
\end{equation}
where
\begin{align*}
\begin{split}
m_1 &= (d_0-d_2) g_0 \\
m_4 &= (d_1-d_3) g_2
\end{split}
\begin{split}
m_2 &= (d_1+d_2) \frac{g_0+g_1+g_2}{2} \\
m_3 &= (d_2-d_1) \frac{g_0-g_1+g_2}{2}
\end{split}
\end{align*}

This algorithm uses just 4 multiplications and is therefore minimal by the formula $\mu (F(2,3)) = 2+3-1=4$. It also uses 4 additions involving the data, 3 additions and 2 multiplications by a constant involving the filter (the sum $g0+g2$ can be computed just once), and 4 additions to reduce the products to the final result.

Fast filtering algorithms can be written in matrix form as:
\begin{equation}\label{eq:F_2_3_matrix}
Y = A^T \big[(G g) \odot( B^T d)\big]
\end{equation}
where $\odot$ indicates element-wise multiplication. For $F(2,3)$, the matrices are:
\begin{equation}\label{eq:F_2_2_3_3_matrices}
\begin{split}
B^T &= \begin{bmatrix*}[r]
1 && 0 && -1 && 0 \\
0 && 1 && 1 && 0 \\
0 && -1 && 1 && 0 \\
0 && 1 && 0 && -1
\end{bmatrix*} \\ 
G  &= \begin{bmatrix*}[r]
1 && 0 && 0 \\
\frac{1}{2} && \frac{1}{2} && \frac{1}{2} \\
\frac{1}{2} && - \frac{1}{2} && \frac{1}{2} \\
0 && 0 && 1
\end{bmatrix*} \\
A^T &= \begin{bmatrix*}[r]
1 && 1 && 1 && 0\\
0 && 1 && -1 && -1 
\end{bmatrix*} \\
g &= \begin{bmatrix*}[r]
g_0 & g_1 & g_2
\end{bmatrix*}^T \\
d &= \begin{bmatrix*}[r]
d_0 & d_1 & d_2 & d_3
\end{bmatrix*}^T
\end{split}
\end{equation}

A minimal 1D algorithm $F(m,r)$  is nested with itself to obtain a minimal 2D algorithm, $F(m \times m, r \times r)$ like so:
\begin{equation}\label{eq:general_2D_convolution}
Y = A^T \bigg[ [G g G^T] \odot [B^T d B] \bigg] A
\end{equation}
where now $g$ is an $r \times r$ filter and $d$ is an $(m+r-1) \times (m+r-1)$ image tile. The nesting technique can be generalized for non-square filters and outputs, $F(m \times n, r \times s)$, by nesting an algorithm for $F(m, r)$ with an algorithm for $F(n, s)$.

$F(2\times 2, 3\times 3)$ uses $4 \times 4 = 16$ multiplications, whereas the standard algorithm uses $2 \times 2 \times 3 \times 3 = 36$. This is an arithmetic complexity reduction of $\frac{36}{16} = 2.25$. The data transform uses 32 additions, the filter transform uses 28 floating point instructions, and the inverse transform uses 24 additions. 

Algorithms for $F(m \times m, r \times r)$ can be used to compute convnet layers with $r \times r$ kernels. Each image channel is divided into tiles of size $(m+r-1)\times (m+r-1)$, with $r-1$ elements of overlap between neighboring tiles, yielding $P = \ceil{H/m} \ceil{W/m}$ tiles per channel, $C$. $F(m \times m, r \times r)$ is then computed for each tile and filter combination in each channel, and the results are summed over all channels. 

Substituting $U = G g G^T$ and $V = B^T d B$, we have:
\begin{equation}\label{eq:general_2D_convolution_UV}
Y = A^T \big[ U \odot V \big] A
\end{equation}

Labeling tile coordinates as $(\widetilde{x},\widetilde{y})$, we rewrite the convnet layer formula (\ref{eq:convnetcorrelation}) for a single image $i$, filter $k$, and tile coordinate $(\widetilde{x},\widetilde{y})$ as:
\begin{equation}\label{eq:fastconvnet}
\begin{split}
Y_{i, k, \widetilde{x},\widetilde{y}} &= \sum\limits_{c=1}^C D_{i,c,\widetilde{x},\widetilde{y}} \ast G_{k,c} \\
 &= \sum\limits_{c=1}^C A^T \bigg[ U_{k,c} \odot V_{c, i,\widetilde{x},\widetilde{y}} \bigg] A \\
 &= A^T \bigg[ \sum\limits_{c=1}^C U_{k,c} \odot V_{c, i,\widetilde{x},\widetilde{y}} \bigg] A
\end{split}
\end{equation}

Thus we can reduce over $C$ channels in transform space, and only then apply the inverse transform $A$ to the sum. This amortizes the cost of the inverse transform over the number of channels.

We examine the sum
\begin{equation}\label{eq:fastconvnet_sum}
M_{k,i,\widetilde{x},\widetilde{y}} =  \sum\limits_{c=1}^C U_{k,c} \odot V_{c,i,\widetilde{x},\widetilde{y}}
\end{equation}
and simplify the notation by collapsing the image/tile coordinates $(i,\widetilde{x},\widetilde{y})$ down to a single dimension, $b$. We also label each component of the element-wise multiplication separately, as $(\xi, \nu)$, yielding:
\begin{equation}
M_{k,b}^{(\xi,\nu)} = \sum\limits_{c=1}^C U^{(\xi,\nu)}_{k,c} V^{(\xi,\nu)}_{c,b}
\end{equation}

This equation is just a matrix multiplication, so we can write:
\begin{equation}\label{eq:fastconvnet_transformdims_individual}
M^{(\xi,\nu)} = U^{(\xi,\nu)} V^{(\xi,\nu)}
\end{equation}

Matrix multiply has efficient implementations on CPU, GPU, and FPGA platforms, owing to its high computational intensity. Thus we have arrived at the practical implementation for the fast algorithm listed in Algorithm~\ref{algo:winograd-fast-filtering}.

\begin{algorithm}
  \caption{Compute Convnet Layer with Winograd Minimal Filtering Algorithm $F(m\times m, r \times r)$}
  \begin{algorithmic}
    \Statex $P=N\lceil H/m \rceil\lceil W/m \rceil$ is the number of image tiles.
    \Statex $\alpha=m+r-1$ is the input tile size.
    \Statex Neighboring tiles overlap by $r-1$.
    \Statex $d_{c,b} \in \mathbb{R}^{\alpha\times \alpha}$ is input tile $b$ in channel $c$.
    \Statex $g_{k,c} \in \mathbb{R}^{r\times r}$ is filter $k$ in channel $c$.
    \Statex $G$, $B^T$, and $A^T$ are filter, data, and inverse transforms.
    \Statex $Y_{k,b} \in \mathbb{R}^{m\times m}$ is output tile $b$ in filter $k$.
    \For  {$k=0$ to $K$} 
    \For  {$c=0$ to $C$} 
    \State        $u = Gg_{k,c}G^T \in \mathbb{R}^{\alpha\times \alpha}$ 
    \State Scatter $u$ to matrices U: $U^{(\xi,\nu)}_{k,c}=u_{\xi,\nu}$
    \EndFor
    \EndFor
	\For {$b=0$ to $P$}
	\For {$c=0$ to $C$}
	\State        $v = B^Td_{c,b}B \in \mathbb{R}^{\alpha\times \alpha} $
    \State Scatter $v$ to matrices V: $V^{(\xi,\nu)}_{c,b}=v_{\xi,\nu}$
    \EndFor 
    \EndFor
    \For {$\xi=0$ to $\alpha$}
    \For {$\nu=0$ to $\alpha$}
	\State $M^{(\xi,\nu)} = U^{(\xi,\nu)} V^{(\xi,\nu)}$
	\EndFor
    \EndFor
    \For  {$k=0$ to $K$} 
	\For {$b=0$ to $P$}
    \State Gather m from matrices M: $m_{\xi,\nu} = M^{(\xi,\nu)}_{k,b}$
	\State      $Y_{k,b} = A^T m A $
	\EndFor
    \EndFor
\end{algorithmic}
\label{algo:winograd-fast-filtering}
\end{algorithm}

Winograd documented a technique for generating the minimal filtering algorithm $F(m,r)$ for any choice of $m$ and $r$. The construction uses the Chinese remainder theorem to produce a minimal algorithm for linear convolution, which is equivalent to polynomial multiplication, then transposes the linear convolution algorithm to yield a minimal filtering algorithm. The reader is referred to Winograd's seminal book~\cite{winograd1980arithmetic}, or Blahut's book~\cite{blahut2010fast} for a modern treatment of the subject. We provide derivations of the specific algorithms used in this paper in the supplementary material.

\subsection{F(3x3,2x2)}

Training a network using stochastic gradient descent requires computation of the gradients with respect to the inputs and weights. For a convnet layer, the gradient with respect to the inputs is a convolution of the next layer's backpropagated error, of dimension $N \times K \times H \times W$, with a flipped version of the layer's $R \times S$ filters. Therefore it can be computed using the same algorithm that is used for forward propagation.

The gradient with respect to the weights is a convolution of the layer inputs with the backpropagated errors, producing $R \times S$ outputs per filter and channel. Therefore we need to compute the convolution $F(R \times S,H\times W)$, which is impractical because $H \times W$ is much too large for our fast algorithms. Instead we decompose this convolution into a direct sum of smaller convolutions, for example  $F(3 \times 3, 2 \times 2)$. Here the algorithm's $4 \times 4$ tiles  are overlapped by 2 pixels in each dimension, and the $3 \times 3$ outputs are summed over all tiles to form $F(3 \times 3, H \times W)$.

The transforms for $F(3 \times 3, 2 \times 2)$ are given by:
\begin{equation}\label{eq:F_3_2_matrices}
\begin{split}
B^T &= \left[
\begin{array}{rrrr}
 1 & 0 & -1 & 0 \\
 0 & 1 & 1 & 0 \\
 0 & -1 & 1 & 0 \\
 0 & -1 & 0 & 1 \\
\end{array}
\right] \\
A^T &= \left[
\begin{array}{rrrr}
 1 & 1 & 1 & 0 \\
 0 & 1 & -1 & 0 \\
 0 & 1 & 1 & 1 \\
\end{array}
\right]
\end{split},
\begin{split}
G &= \left[
\begin{array}{rr}
 1 & 0 \\
 \frac{1}{2} & \frac{1}{2} \\
 \frac{1}{2} & -\frac{1}{2} \\
 0 & 1 \\
\end{array}
\right] \\
\end{split}
\end{equation}

With $(3+2-1)^2$ = 16 multiplies versus direct convolution's $3 \times 3 \times 2 \times 2 = 36$ multiplies, it achieves the same $36/16 = 2.25$ arithmetic complexity reduction as the corresponding forward propagation algorithm. 

\subsection{F(4x4,3x3)}

A minimal algorithm for  $F(4, 3)$ has the form:
\begin{equation}\label{eq:F_4_3_matrices}
\begin{split}
B^T &= \left[
\begin{array}{rrrrrr}
 4 & 0 & -5 & 0 & 1 & 0 \\
 0 & -4 & -4 & 1 & 1 & 0 \\
 0 & 4 & -4 & -1 & 1 & 0 \\
 0 & -2 & -1 & 2 & 1 & 0 \\
 0 & 2 & -1 & -2 & 1 & 0 \\
 0 & 4 & 0 & -5 & 0 & 1 \\
\end{array}
\right] \\
G &= \left[
\begin{array}{rrr}
 \frac{1}{4} & 0 & 0 \\
 -\frac{1}{6} & -\frac{1}{6} & -\frac{1}{6} \\
 -\frac{1}{6} & \frac{1}{6} & -\frac{1}{6} \\
 \frac{1}{24} & \frac{1}{12} & \frac{1}{6} \\
 \frac{1}{24} & -\frac{1}{12} & \frac{1}{6} \\
 0 & 0 & 1 \\
\end{array}
\right] \\
A^T &= \left[
\begin{array}{rrrrrr}
 1 & 1 & 1 & 1 & 1 & 0 \\
 0 & 1 & -1 & 2 & -2 & 0 \\
 0 & 1 & 1 & 4 & 4 & 0 \\
 0 & 1 & -1 & 8 & -8 & 1 \\
\end{array}
\right]
\end{split}
\end{equation}

The data transform uses 13 floating point instructions, the filter transform uses 8, and the inverse transform uses 10.

Applying the nesting formula yields a minimal algorithm for $F(4 \times 4, 3 \times 3)$ that uses $6 \times 6 = 36$ multiplies, while the standard algorithm uses $4 \times 4 \times 3 \times 3 = 144$. This is an arithmetic complexity reduction of $4$.

The 2D data transform uses $13(6+6) = 156$ floating point instructions, the filter transform uses $8(3+6) = 72$, and the inverse transform uses $10(6+4)=100$.

The number of additions and constant multiplications required by the minimal Winograd transforms increases quadratically with the tile size~\cite[p.~211]{madisetti2010digital}. Thus for large tiles, the complexity of the transforms will overwhelm any savings in the number of multiplications.

The magnitude of the transform matrix elements also increases with increasing tile size. This effectively reduces the numeric accuracy of the computation, so that for large tiles, the transforms cannot be computed accurately~\cite[p.~28]{winograd1980arithmetic}.

Convnets require surprisingly little numeric precision~\cite{DBLP:journals/corr/CourbariauxBD14,gupta2015deep}. This implies that we can sacrifice some numeric accuracy in the filtering computation without affecting the accuracy of the convnet. We examine the possibility of $F(6 \times 6, 3 \times 3)$ in the supplementary material.

\subsection{Fast Fourier Transform}

The Fast Fourier Transform (FFT) can be used to produce a tiled convolution algorithm that has the same form as Algorithm ~\ref{algo:winograd-fast-filtering}. The main difference is that the transform matrices are replaced with FFT and inverse FFT, and point-wise multiplication of complex FFT components yields cyclic convolution. Only $m \times n$ components of the $(m+r-1) \times (n+s-1)$ cyclic convolution are valid, the rest must be discarded, and the tiles must be overlapped  by $r-1$ and $s-1$ in order to recompute the discarded outputs. This technique is referred to as overlap and save~\cite[p.~195]{blahut2010fast}.

The similarity of overlap and save to our approach makes for an easy comparison. With FFT based convolution, the multiply stage still uses 1 multiply per input, but now the operands are complex numbers. Direct multiplication of complex numbers requires 4 real multiplications. Thankfully, a couple of tricks reduce the complexity further.

The Fourier transform of a real signal has Hermitian symmetry, which reduces the number of unique products in each $U \odot V$ by almost half. FFT based convnet implementations have exploited this property~\cite{DBLP:journals/corr/MathieuHL13,DBLP:journals/corr/VasilacheJMCPL14}. Specifically, the discrete Fourier transform of a $\alpha\times \alpha$ array of real values can be represented with an array of $\alpha \times (\lfloor \frac{\alpha}{2} \rfloor + 1)$ complex values. Furthermore $U^H V^H = (UV)^H$, so the products of the missing values can be reconstructed simply by taking the complex conjugate of the computed values. Thus the multiply stage of the FFT convnet algorithm with tile size $\alpha = m + r - 1$ requires $N \lceil \frac{H}{m} \rceil \lceil \frac{W}{m} \rceil C K \alpha (\lfloor \frac{\alpha}{2} \rfloor + 1)$ complex multiplications, or $(\lfloor \frac{\alpha}{2} \rfloor + 1)/\alpha$ complex multiplies per input. 

Using the standard algorithm for multiplying complex numbers, this equals $4 (\lfloor \frac{\alpha}{2} \rfloor + 1)/\alpha > 2$ real multiplies per input.

Another technique, which to our knowledge has not been used in convnets, is to use a fast algorithm to multiply complex numbers with 3 real multiplications~\cite{winograd1980arithmetic}:
\begin{equation}
\begin{split}
(x_0 + i x_1)(y_0 + i y_1) &= \left[x_0 y_0 - x_1 y_1, i(x_0 y_1 + x_1 y_0)\right] \\
&= \left[u_c v_a + u_a v_c, i (u_a v_c - u_b v_b)\right]
\end{split}
\end{equation}
where
\begin{equation}
\begin{split}
u_a &= x_0 \\
u_b &= x_0 + x_1 \\
u_c &= x_1 - x_0
\end{split},
\begin{split}
v_a &= y_0 \\
v_b &= y_1 \\
v_c &= y_0 + y_1
\end{split}
\end{equation}

An FFT based convnet algorithm can incorporate this by modifying the FFT transforms of the filter and data to output the the real valued matrices $(U_a, U_b, U_c)$ and $(V_a, V_b, V_c)$ instead of the complex valued matrices $U$ and $V$. This adds 2 floating point instructions per output to the filter transform, and 1 to the data transform. It also increases the memory footprint of each matrix by half.

Then we can calculate $M = U V$ using 3 calls to a standard real matrix multiply function (\eg SGEMM):
\begin{equation}
\begin{split}
T &= U_a V_c \\
M_1 &= -U_b V_b + T,
\end{split}
\begin{split}
M_0 &= U_c V_a + T \\
M &= (M_0, i M_1) 
\end{split}
\end{equation}

The accumulation of temporary matrix $T$ is performed using regular SGEMM with $\beta=1$ and $C=T$, at the cost of adding 2 floating point instructions per output. We can think of these instructions as adding to the inverse transform cost. The temporary matrix $T$ increases memory use by half, so that the total workspace size is approximately twice that of FFT based convolution with direct CGEMM.

Combining Hermitian symmetry with fast CGEMM gives us a multiplication stage with $3 (\lfloor \frac{\alpha}{2} \rfloor + 1)/\alpha > 1.5$ real multiplies per input. Recall that the multiply stage of the Winograd algorithms is always 1 real multiply per input. Thus even with fast CGEMM, FFT base convolution must use a significantly larger tile size in order to rival the arithmetic complexity of the Winograd algorithms.

For the FFT transform itself, we consider the split-radix FFT algorithm, which is the minimal practical FFT algorithm when $N$ is a power of 2~\cite[p.~150]{madisetti2010digital}. We assume the 2D FFT transform is constructed using row-column composition, and borrow the complexity figures from the DSP Handbook~\cite[pp.~173,175]{madisetti2010digital} for Table \ref{table:complexity-versus-tile-size}.

\begin{table}
\begin{center}
\begin{tabular}{|c|cccc|c r|}
\hline
\multirow{2}{*}{Tile} & \multicolumn{4}{c|}{Winograd} &  \multicolumn{2}{c|}{FFT} \\
\cline{2-7}
 & $\alpha'$ & $\beta'$ & $\gamma'$    & $\delta'$ & $\alpha'$ & $\beta',\gamma',\delta'$ \\
\hline
\hline
3 &  9.00 & \multicolumn{1}{c}{-} & \multicolumn{1}{c}{-} & \multicolumn{1}{c|}{-}  & & \\
4 & 4.00 & 2.00 & 1.75 & 1.50 & & \\
5 & 2.78 & 3.60 & 2.24 & 2.24 & &  \\ 
\textbf{6} &  \textbf{2.25} & \textbf{4.33} & \textbf{2.00} & \textbf{2.78}  & & \\ 
8 & 1.78 & 6.50 & 2.23 & 4.38 & 4.44 & 2.42 \\
16 &  &  &  &  & 2.94 & 4.23 \\
32 &  &  &  &   & 2.42 & 6.24 \\
\textbf{64} &  &  &  &   & \textbf{2.20} & \textbf{8.30} \\
128 &  &  &  &  & 2.10 & 10.37  \\
256 &  &  &  &   & 2.05 & 12.42 \\
\hline
\end{tabular}
\end{center}
\caption{Multiply ($\alpha'$), data transform ($\beta'$), filter transform ($\gamma'$), and inverse transform ($\delta'$) normalized arithmetic complexity versus tile size, for both Winograd and FFT based convolution. F(4x4,3x3) has tile size 6. Direct convolutions has tile size 3.}
\label{table:complexity-versus-tile-size}
\end{table}

\begin{table}
\begin{center}
\begin{tabular}{|c|c r r r|}
\hline
\multirow{2}{*}{Tile}& \multicolumn{4}{c|}{FFT with Fast CGEMM}\\
\cline{2-5}
 & $\alpha'$ & $\beta'$ & $\gamma'$ & $\delta'$ \\
\hline
\hline
8 & 3.33 & 3.77 & 4.30 & 4.30 \\
\textbf{16}  & \textbf{2.20} & \textbf{6.23} & \textbf{6.82} & \textbf{6.82} \\
32 & 1.81 & 8.94 & 9.57 & 9.57 \\
64 & 1.65 & 11.72 & 12.36 & 12.36 \\
128 & 1.57 & 14.48 & 15.14 & 15.14 \\
256 & 1.54 & 17.22 & 17.88 & 17.88 \\
\hline
\end{tabular}
\end{center}
\caption{Normalized arithmetic complexity for FFT filtering with fast CGEMM. Fast CGEMM uses 3 multiplies per complex multiply instead of 4, but has slightly greater transform overhead and uses more memory.}
\label{table:complexity-versus-tile-size-fast-cgemm}
\end{table}

\section{Arithmetic Complexity Analysis}

In our model of fast convnets, the arithmetic complexity of the multiplication stage is:
\begin{equation}
M = N \lceil H/m \rceil \lceil W/n \rceil CK(m+R-1)(n+S-1)
\end{equation}

When $m=n=1$, the formula equals the arithmetic complexity of direct convolution. Therefore direct convolution is the minimal algorithm for $F(1 \times 1, R \times S)$

Although our analysis employs minimal convolutions, the convnet layer itself is still not minimal because it  performs more convolutions than are strictly necessary. We could reduce the number of convolutions by employing Strassen recursions as in~\cite{cong2014minimizing}, but each recursion reduces all 3 dimensions of our matrices by half while providing only an $\frac{8}{7}$ reduction in arithmetic complexity. The matrix multiplications cannot be computed efficiently if $C$ or $K$ is too small. Fast convolution alone provides a $2.25$ or larger arithmetic complexity reduction while shrinking only the largest dimension of the matrix, $P$. Still, for layers with large $C$, $K$, and $P$, it may be worthwhile to perform Strassen recursions in addition to fast convolution. We leave this as an area for further research.
 
In order to simplify the equations, we will henceforth assume that $W/m$ and $H/n$ have no remainders. We also assume square filters and blocks, $R=S$ and $m=n$.
 
The multiplication complexity can be rewritten as:
\begin{equation}
\begin{split}
M &= (m+R-1)^2/m^2 NHWCK \\
&= \alpha' NHWCK 
\end{split}
\end{equation}
where $\alpha=(m+R-1)^2$ and $\alpha' =\alpha/m^2$
 
The total arithmetic complexities of the data, filter, and inverse transforms can be written as:
\begin{equation}
\begin{split}
T(D) &= \beta/m^2 NHWC \\
T(F) &= \gamma CK \\
T(I) &= \delta/m^2  NHWK
\end{split}
\end{equation}
where $\beta$, $\gamma$, and $\delta$ are the number of floating point instructions used by the corresponding transforms for single tiles.
 
Dividing the complexity of each transform by $M$ yields its relative complexity:
\begin{equation}
\begin{split}
T(D)/M &= \beta/(K \alpha^2) = \beta'/K \\
T(F)/M &= \gamma/(NHW\alpha^2/m^2) \\ 
       &= \gamma/(P\alpha^2) = \gamma'/P \\
T(I)/M &= \delta/(C\alpha^2) = \delta'/C
\end{split}
\end{equation}
 
We call $\beta'$, $\gamma'$, and $\delta'$ the normalized arithmetic complexities of the data, filter, and inverse transforms, respectively. $P=NHW/m^2$ is the number of tiles per channel.
 
Adding the terms for each stage gives the total arithmetic complexity of the convnet layer:
\begin{equation}
L = \alpha' (1 + \beta'/K + \gamma'/P + \delta'/C) NHWCK
\label{eq:totalcomplexity}
\end{equation}
  
In order to achieve a large speedup, the multiplication complexity $\alpha'$ must be as small as possible, and the transform complexities $\beta'$, $\gamma'$, and $\delta'$ must each be small compared with $K$, $P$, and $C$, respectively.
 
For direct convolution, $\alpha' = \alpha^2 = R^2$ and $\beta'=\gamma'=\delta'=0$. Therefore the maximum speedup of a fast algorithm versus direct convolution is $R^2/\alpha'$.

We list the normalized transform complexity for different tile sizes and algorithms in Tables \ref{table:complexity-versus-tile-size} and \ref{table:complexity-versus-tile-size-fast-cgemm}. Due to its similarity to our approach, FFT based convolution complexity can also be measured with Equation~\ref{eq:totalcomplexity}.

FFT based convnet layers with direct CGEMM must use tile size at least $64 \times 64$ to equal the multiplication stage complexity of Winograd $F(4 \times 4, 3 \times 3)$ and its $6 \times 6$ tile, but then it incurs much greater transform overhead. Also a $64 \times 64$ tile will waste computation on many unwanted pixels for images with sizes that are not close to a multiple of $62 \times 62$. Even for moderate size layers, a moderate to large minibatch must be used, or there will be too few tiles to compute the CGEMM efficiently. Finally, the memory used by a single transformed filter channel is $64 \times 64 = 4096$ units, which is a large expansion of the $3 \times 3 = 9$ unit filter. The 6x6 tile of $F(4 \times 4)$ expands the same filter to $6 \times 6 = 36$ units.

FFT based convnet layers with fast CGEMM can be much more competitive with Winograd algorithms. They have multiplication stage parity with tile size $16$, and reasonable transform complexity. Also tile size $16$ generates a reasonably large number of tiles with large convnet layers or moderate batch size.

Even with fast CGEMM, the larger tile size compared to Winograd means FFT based convnet implementations must have a large memory workspace to hold transformed data. A decent amount of transformed data must be held in order to amortize transform cost and to generate matrices with large enough dimensions so that the multiply stage is efficient. This is problematic for current GPUs, which have a limited amount of on chip memory. CPUs have large caches and might therefore compute FFT based convolution more efficiently.

\section{GPU Implementation}

We implemented $F(2\times 2, 3 \times 3)$ for NVIDIA Maxwell GPUs and tested on the NVIDIA Titan X model. 

The small $4\times 4$ tile size and light weight transforms of $F(2 \times 2, 3 \times 3)$ make possible a fused implementation of the algorithm stages, where the the data and filter transform, 16 batched matrix multiplies (GEMMs), and inverse transform are all computed in the same block. Another resource limit is the instruction cache, which can only fit about 720 instructions. Our main loop is larger than this, but aligning the start of the loop with the 128 byte instruction cache-line boundary helps mitigate the cost of a cache miss.

The 16 batched GEMMs compute $32 \times 32$ outputs, which enables us to fit the workspace in the registers and shared memory of a single block and still have 2 active blocks per SM for latency hiding. Zero padding is implicit through use of predicates. If the predicate deselects a global image load, the zero value is loaded with a dual issued I2I instruction.

Image data is stored in CHWN order to facilitate contiguous and aligned memory loads, significantly reducing over-fetch. We employ a ``super blocking'' strategy to load 32 tiles of size $4 \times 4$ from a configurable number of images, rows, and columns. For $N>=32$, we load tiles from 32 separate images. For $N<32$, we load a super block of $X\times Y=32/N$ tiles per image. This strategy facilitates efficient loads with small batch sizes, as the $W\times N$ dimensions of the input data are contiguous in memory. Furthermore, the 2 pixel overlap between adjacent tiles causes high L1 cache hit rates when using several tiles in a super block. 

We also employ L2 cache blocking to increase the re-use of overlapping blocks. Since the number of image tiles is typically much larger than the number of filters, our block mapping iterates over a group of up to 128 filters in the inner loop, and then iterates over all image tiles in the second loop. All channels of the filter group fit in L2 cache, so each filter will only be loaded once from DDR memory, and each image tile will be loaded $\lceil K/128 \rceil$ times as we iterate over the filter groups. This strategy reduces DDR memory bandwidth by almost half.

We implemented a version of our kernel that loads fp16 data, which decreases global memory bandwidth. We also implemented a variant that we call ``FX'' that runs a filter transform kernel first and stores the result in a workspace buffer. The convolution kernel loads transformed filter values from the workspace as needed. The size of the workspace is only $16KC$ units of memory, which equals just 16MB when $K=C=512$ and data is fp32.

\section{Experiments}

We ran accuracy and speed experiments with VGG Network E~\cite{simonyan2014very}. This is a  deep network that uses $3 \times 3$ filters exclusively in the convolution layers, which are summarized in Table \ref{table:vgg}.

\begin{table}
\begin{center}
\begin{tabular}{ |l| c c c S[table-format=3.2]| }
\hline
Layer & Depth & $C \times H \times W$ & K & GFLOPs \\
\hline
\hline
conv 1.1 & 1 & $3 \times 224 \times 224 $  & $64$ & 0.17 \\
conv 1.2 & 1 & $64 \times 224 \times 224 $  & $64$ & 3.70 \\
conv 2.1 & 1 & $64 \times 112 \times 112 $  & $128$ & 1.85 \\
conv 2.2 & 1 & $128 \times 112 \times 112 $  & $128$ & 3.70 \\
conv 3.1 & 1 & $128 \times 56 \times 56 $  & $256$ & 1.85 \\
conv 3.2 & 3 & $256 \times 56 \times 56 $  & $256$ & 11.10 \\
conv 4.1 & 1 & $256 \times 28 \times 28 $  & $512$ & 1.85 \\
conv 4.2 & 3 & $512 \times 28 \times 28 $  & $512$ & 11.10 \\
conv 5 & 4 & $512 \times 14 \times 14 $  & $512$ & 3.70 \\
\hline
\hline
Total & & & & 39.02 \\ 
\hline
\end{tabular}
\end{center}
\caption{Convolution layers of VGG network E. All layers uses $3 \times 3$ filters. Depth indicates the number of times a given layer shape occurs in the network. GFLOPs is weighted by depth and assumes N=1.}
\label{table:vgg}
\end{table}

We tested the accuracy of our fast algorithms with both single precision (fp32) and half precision (fp16) data and filters. In all tests we used fp32 arithmetic instructions. We used random data and filters from the uniform distribution $\left[ -1,1 \right]$ and measured absolute element error. Ground truth was computed by direct convolution using a double precision accumulator for reductions.

We measured the speed of our GPU implementation of $F(2\times 2, 3\times 3)$ and compared with cuDNN v3~\cite{cuDNN} on a superclocked NVIDIA Titan X GPU. We disabled boost clock and observed a maximum clock rate of 1126MHz. The GPU has 3072 cores, yielding a device peak throughput of $2 \times 3072 \times 1126= 6.96$ TFLOPS.

Speed for a given layer was calculated by dividing the number of GFLOPs of computation required by direct convolution, as tabulated in \ref{table:vgg}, by the run time in milliseconds to yield Effective TFLOPS. The reduction of arithmetic complexity allows fast algorithms to have Effective TFLOPS that can exceed device peak throughput.

Total GFLOPs and run time were calculated by weighting the GFLOPs and run time for each layer by its depth, and total throughput was calculated as the ratio of the two.

\section{Results}

Table \ref{table:fp32-accuracy} shows the numeric accuracy of the different convolution layer algorithms tested with single precision (fp32) and half precision (fp16) input data and filters.

$F(2\times 2, 3 \times 3)$ is actually slightly more accurate than direct convolution. Its simple transforms do not lose much precision, and its multiplication stage performs a reduction over $C$ channels, rather than the $R S C$ filter elements reduced by direct convolution. $F(4\times 4, 3 \times 3)$ has a larger error, but it is still more accurate than direct convolution with fp16 data.

All tested algorithms are equally accurate with fp16 data. Here accuracy is limited by the precision of the inputs. Because direct convolution is  accurate enough for training and inference with low precision data~\cite{DBLP:journals/corr/CourbariauxBD14,gupta2015deep}, we conclude that $F(4\times 4, 3 \times 3)$ is too.

Table~\ref{table:speed} and Table ~\ref{table:speed-fp16} show the total throughput for VGG Network E layers for cuDNN and our $F(2\times 2, 3\times 3)$ implementation for fp32 and fp16 data for different batch sizes. 

For fp32 data, $F(2\times 2, 3\times 3)$ is $1.48$X at $N=64$ and $2.26$X as fast at $N=1$. The throughput at $N=16$ is $9.49$ TFLOPS. For fp16 data, $F(2\times 2, 3\times 3)$ extends its lead over cuDNN, recording $10.28$ TFLOPS throughput for $N=64$. $N=8$ performance is still very good at $9.57$ TFLOPS.

\begin{figure}[t]
\begin{center}
   \includegraphics[width=1.0\linewidth]{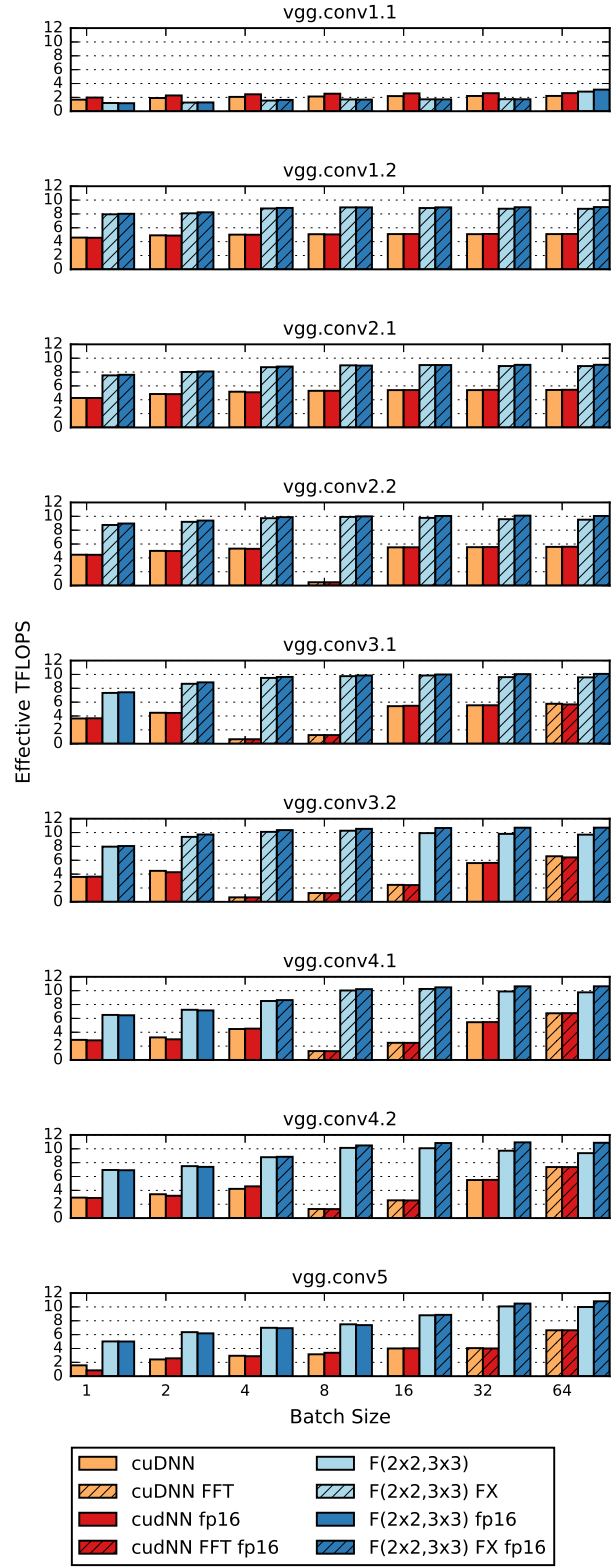}
\end{center}
\caption{VGG net Effective TFLOPS vs. batch size for cuDNN and $F(2\times 2, 3\times 3)$ on a $6.96$ TFLOPS NVIDIA Titan X GPU.}
\label{fig:throughput}
\end{figure}

Figure~\ref{fig:throughput} shows throughput by layer. Hatch marks indicate the layers where cuDNN used the FFT algorithm, otherwise direct convolution was used. For $F(2\times 2, 3\times 3)$, hatch marks indicate that the external filter transform (FX) was used, otherwise the fused transform was faster.

cuDNN appears to erroneously select its FFT algorithm for intermediate values of $N$ despite the fact that it performs very poorly, under $2$ TFLOPS. While this is probably just a bug, it is revealing. Low performance at moderate values of $N$  suggests that the FFT convolution implementation either uses large tiles, or possibly just a single tile per image, as in~\cite{DBLP:journals/corr/VasilacheJMCPL14}, which leads to inefficient multiplication stages unless $N$ is large. At large $N$, cuDNN FFT performs much better, but stays well under $8$ TFLOPS.

$F(2\times 2, 3\times 3)$ performs better than cuDNN at every layer and batch size, except layer conv1.1, which contributes less than $0.5\%$ of the total network computation.

In general, we found that the FX variant of our implementation performed best unless the number of filters and channels was very large. Computing the filter transform is heavily memory bound, therefore transforming a larger filter bank decreases computational efficiency.

The worst $F(2\times 2, 3\times 3)$ performance occurs for the $14 \times 14$ layers when $N=1$. In this case the $8\times 4$ superblock runs over the image boundary and computes unwanted pixels. Throughput on this layer configuration is still over $5$ TFLOPS, where cuDNN performance is just $1.6$ TFLOPS.

cuDNN FFT uses a global memory workspace up to $2.6$ GB in our experiments. By contrast, our fused $F(2 \times 2, 3 \times 3)$ implementation does not use any global workspace, and the FX variant uses no more than $16$ MB.

$F(2\times 2, 3\times 3)$ performance shows new capabilities for high throughput and small batch size with state of the art convolutional neural networks. We expect performance to increase again when $F(4\times 4, 3\times 3)$ is implemented.

\begin{table}
\begin{center}
\begin{tabular}{ |c|c c c|c| }
\hline
\multirow{2}{*}{Layer} & \multicolumn{3}{c|}{fp32} & \multirow{2}{*}{fp16}\\
 & Direct & F(2x2,3x3) & F(4x4,3x3) & \\
\hline
\hline
1.2 & 4.01E-05 & 1.53E-05 & 2.84E-04 & 1.14E-02\\
2.2 & 8.01E-05 & 2.86E-05 & 5.41E-04 & 1.45E-02\\
3.2 & 1.53E-04 & 5.34E-05 & 9.06E-04 & 1.99E-02\\
4.2 & 3.20E-04 & 5.34E-05 & 1.04E-03 & 3.17E-02\\
5 & 3.43E-04 & 4.20E-05 & 1.08E-03 & 2.61E-02\\
\hline
\end{tabular}
\end{center}
\caption{Maximum element error on VGG network layers. With fp32 data, $F(2 \times 2,3 \times 3)$ is more accurate than direct convolution. With fp16 data, all algorithms are equally accurate.}
\label{table:fp32-accuracy}
\end{table}

\begin{table}
\begin{center}
\begin{tabular}{|r | r r | r r | r|}
\hline
\multirow{2}{*}{N} & \multicolumn{2}{c|}{cuDNN} & \multicolumn{2}{c|}{F(2x2,3x3)} & \multirow{2}{*}{Speedup}\\
 & msec & TFLOPS & msec & TFLOPS & \\
\hline
\hline
  1 &   12.52 &     3.12 &     5.55 &     7.03 &     2.26X \\
  2 &   20.36 &     3.83 &     9.89 &     7.89 &     2.06X \\
  4 &  104.70 &     1.49 &    17.72 &     8.81 &     5.91X \\
  8 &  241.21 &     1.29 &    33.11 &     9.43 &     7.28X \\
 16 &  203.09 &     3.07 &    65.79 &     9.49 &     3.09X \\
 32 &  237.05 &     5.27 &   132.36 &     9.43 &     1.79X \\
 64 &  394.05 &     6.34 &   266.48 &     9.37 &     1.48X \\
\hline
\end{tabular}
\end{center}
\caption{cuDNN versus $F(2\times 2, 3\times 3)$ performance on VGG Network E with fp32 data. Throughput is measured in Effective TFLOPS, the ratio of direct algorithm GFLOPs to run time.}
\label{table:speed}
\end{table}

\begin{table}
\begin{center}
\begin{tabular}{|r | r r | r r | r|}
\hline
\multirow{2}{*}{N} & \multicolumn{2}{c|}{cuDNN} & \multicolumn{2}{c|}{F(2x2,3x3)} & \multirow{2}{*}{Speedup}\\
 & msec & TFLOPS & msec & TFLOPS & \\
\hline
\hline
  1 &   14.58 &     2.68 &     5.53 &     7.06 &     2.64X \\
  2 &   20.94 &     3.73 &     9.83 &     7.94 &     2.13X \\
  4 &  104.19 &     1.50 &    17.50 &     8.92 &     5.95X \\
  8 &  241.87 &     1.29 &    32.61 &     9.57 &     7.42X \\
 16 &  204.01 &     3.06 &    62.93 &     9.92 &     3.24X \\
 32 &  236.13 &     5.29 &   123.12 &    10.14 &     1.92X \\
 64 &  395.93 &     6.31 &   242.98 &    10.28 &     1.63X \\
\hline
\end{tabular}
\end{center}
\caption{cuDNN versus $F(2\times 2, 3\times 3)$ performance on VGG Network E with fp16 data.}
\label{table:speed-fp16}
\end{table}

\clearpage

\bibliographystyle{plain}
\bibliography{mybib.bib}

\end{document}